\documentclass[nohyperref]{article}

\usepackage{microtype}
\usepackage{graphicx}
\usepackage{subfigure}
\usepackage{booktabs} 

\usepackage{hyperref}

\usepackage[accepted]{icml2022}

\usepackage{amsmath}
\usepackage{amssymb}
\usepackage{mathtools}
\usepackage{amsthm}

\usepackage{algorithm}
\usepackage{algorithmic}

\usepackage[capitalize,noabbrev]{cleveref}

\theoremstyle{plain}

\theoremstyle{definition}

\theoremstyle{remark}

\usepackage[textsize=tiny]{todonotes}

\icmltitlerunning{Latent Variable Models for Bayesian Causal Discovery}

\begin{document}

\twocolumn[
\icmltitle{Latent Variable Models for Bayesian Causal Discovery}



\icmlsetsymbol{equal}{*}

\begin{icmlauthorlist}
\icmlauthor{Jithendaraa Subramanian}{mila,mcgill}
\icmlauthor{Yashas Annadani}{kth}
\icmlauthor{Ivaxi Sheth}{mila,ets}
\icmlauthor{Stefan Bauer\textsuperscript{*}}{kth}
\icmlauthor{Derek Nowrouzezahrai}{mila,mcgill}
\icmlauthor{Samira Ebrahimi Kahou}{mila,ets,cifar}
\end{icmlauthorlist}

\icmlaffiliation{mila}{Mila - Québec AI Institute}
\icmlaffiliation{mcgill}{McGill University, Montréal}
\icmlaffiliation{ets}{ÉTS Montréal}
\icmlaffiliation{kth}{KTH, Stockholm}
\icmlaffiliation{cifar}{CIFAR AI Chair}

\icmlcorrespondingauthor{Jithendaraa Subramanian}{jithen.subra@gmail.com}

\icmlkeywords{Machine Learning, Causal Learning, representation learning, latent variable models}

\vskip 0.3in
]



\printAffiliationsAndNotice{\icmlEqualContribution} 

\begin{abstract}
Learning predictors that do not rely on spurious correlations involves building causal representations. However, learning such representations is very challenging. We, therefore, formulate the problem of learning causal representations from high dimensional data and study causal recovery with synthetic data. This work introduces a latent variable decoder model, Decoder BCD, for Bayesian causal discovery and performs experiments in mildly supervised and unsupervised settings. We present a series of synthetic experiments to characterize important factors for causal discovery and show that using known intervention targets as labels helps in unsupervised Bayesian inference over structure and parameters of linear Gaussian additive noise latent structural causal models.
\end{abstract}

\section{Introduction}
Exploiting structure in the data to infer latent variables and capture causal mechanisms is crucial for causal representation learning \cite{tcrl}. Such a representation would allow for counterfactual reasoning in a manner similar to that of humans, thereby moving away from models that rely on exploiting spurious correlations for prediction. 

Causal mechanisms are usually modelled as Bayesian Networks or Directed Acyclic Graphs (DAG) and given information about the causal variables, one can learn the DAG with structure learning algorithms. Recently, there has been a flurry of works advancing structure learning algorithms \cite{lingam, notears, graphvae, dynotears, dibs, vcn, golem, bcdnets, dag_gflow} that learn the structure of a DAG given data samples (of causal variables). Most of these works cast the discrete optimization of learning a DAG into a continuous one that is optimized through gradient descent, thereby sidestepping the computational intractability arising from the super-exponential nature of DAG search in the discrete case. However, all approaches learn a causal DAG on the premise that one has full access to the true causal variables which might not be realistic. A more realistic assumption would be that we have partial or no access to true causal variables and that one has to infer the structure \textit{along with} the causal variables. 

Here, we introduce a fully differentiable latent variable model, Decoder BCD, to study the problem of Bayesian structure learning in linear Gaussian additive noise models, from high dimensional data. We perform synthetic experiments to analyze why unsupervised causal discovery in latent variable models is difficult. Section 2 explains preliminaries for the setup and section 3 gives the problem setup. In section 4, we introduce Decoder BCD, a decoder model for Bayesian Causal Discovery in the latent space before discussing experiments and our findings in section 5. We discuss related work in section 6 before concluding in section 7. 

\begin{figure*}
  \centering
  \includegraphics[height=8cm]{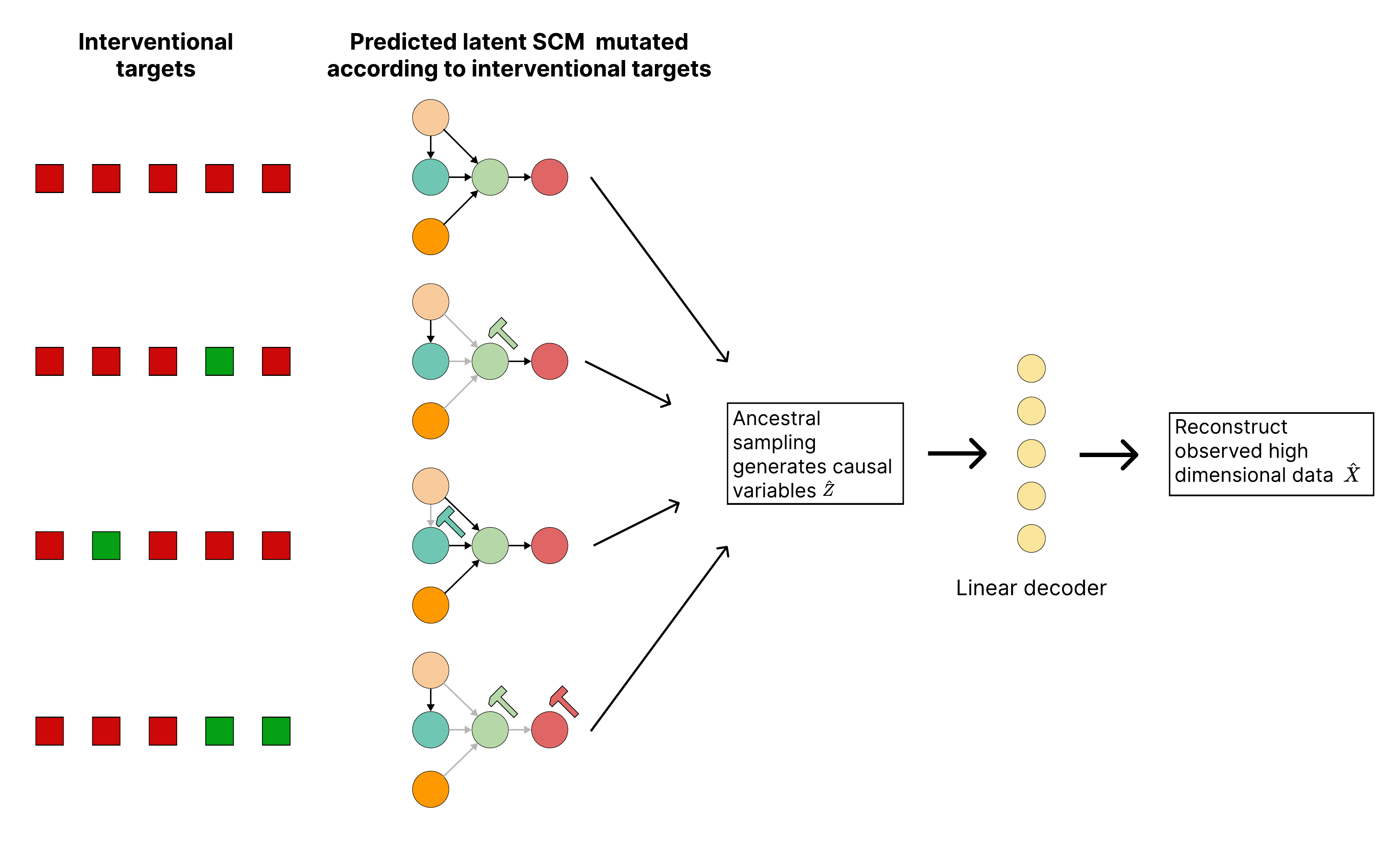} \label{model_img}
  \caption{An illustration of the latent variable (decoder) model for Bayesian causal discovery}
\end{figure*}

\section{Preliminaries}

\textbf{Structural Causal Models (SCM)}: We operate in the framework of SCM \cite{pearl} where node $Z_i$ represents a random causal variable with an independent noise variable $\epsilon_i \sim \mathcal{N}(0, \sigma_i^2)$, and parents $Pa_G(Z_i)$ corresponding to a DAG $G$. We focus primarily on the family of linear Gaussian additive noise models. However, for a DAG to be identifiable from data, one either has to observe a non-Gaussian setting, or, in the case of a Gaussian one, have an equal noise variance assumption (all $\sigma_i = \sigma$) \cite{peters_eq_noi_var}. Since we are in the Gaussian setting, we assume the latter. Thus, we have $d$ causal variables $Z = [Z_1, ... Z_d]$, and a joint distribution entailed by the DAG $G$ such that,
\begin{equation} \label{eq1}
P(Z_1, ... Z_d) = \prod_{i=1}^{d} {P(Z_i | Pa_G(Z_i)) } 
\end{equation}
and the exact values of the $d$ random variables are given by $ z_i = f(Pa_G(z_i)) + \epsilon_i$, where $f(.)$ is a linear function. In our case, $f$ is the weighted sum of values taken on by random variables $Pa_G(Z_i)$, the weights given by the weighted adjacency matrix $W$, such that  $z = W^Tz + \epsilon$. \\

\textbf{Bayesian Causal Discovery Nets (BCD Nets)}: We build Decoder BCD, our latent variable decoder model, upon BCD Nets \cite{bcdnets}. Given samples of true causal variables, BCD Nets is a Bayesian structure learning method that obtains a posterior distribution over causal structures that best explains the data. Similar to other structure learning works \cite{dibs}, BCD Nets introduces the problem of structure learning as a continuous constrained optimization problem. However, the DAG is parameterized such that one always ends up with a DAG and is therefore a hard constraint (in contrast to DiBS which has a soft DAG constraint). This is achieved by formulating the weighted adjacency matrix as $W = {(P L P^T)}^T$ where $P$ is a permutation matrix and $L$ is strictly lower triangular. When $P$ is identity, this is equivalent to having a DAG of fixed ordering with each node $j$ having its possible parents only in the node range $[j+1, d]$. $P$ allows one to transition between node orderings by permuting the rows and columns of $L$. Apart from estimating $W$ (via $P$ and $L$), BCD Nets also infers $\Sigma$, the noise covariance, for the noise variables on each node in the DAG. \\

Thus, overall, BCD Nets formulates the Bayesian Structure Learning problem as inference of $P, L,$ and $\Sigma$ with a unique factorization of the posterior as $q_{\phi}(P, L, \Sigma)$ = $q_{\phi}(P | L, \Sigma) \cdot q_{\phi}(L, \Sigma)$. The model is trained on an ELBO loss (eq. \ref{bcd_elbo}) with a horseshoe prior on $L$, Gumbel Sinkhorn prior \cite{gs} on $P$, and a Gaussian prior on $\Sigma$.
\begin{equation} \label{bcd_elbo}
    \begin{aligned}
        \mathbb{E}_{(L, \Sigma) \sim q_{\phi}} \bigl[ & \mathbb{E}_{P \sim q_{\phi}(. |L, \Sigma)}  \bigl[ \log p(X | P, L, \Sigma) \\ 
        - & \log \frac{q_{\phi}(P | L, \Sigma)}{p(P | L, \Sigma)} \bigr] - \log \frac{q_{\phi}(L, \Sigma)}{p(L, \Sigma)} \bigr]
    \end{aligned}
\end{equation} 
For finer details, we refer the reader to the original work \cite{bcdnets}. Our work focuses more on \emph{extending} BCD Nets to the high dimensional setting and studying unsupervised graph recovery in the latent space.

\section{The Problem Setup}

This work revolves around the Bayesian inference of causal variables $Z$ and the causal structure $G$. Given $n$ samples of high dimensional data $X \in \mathbb{R}^{n \times D}$, we wish to recover a distribution over graph structures $G$ -- the (weighted) adjacency matrix -- and the causal variables $Z \in \mathbb{R}^{n \times d}$. Our setup revolves mostly around the recovery in linear isotropic Gaussian additive noise SCM, which is identifiable. Models like DiBS \cite{dibs}, VCN \cite{vcn}, and BCD Nets \cite{bcdnets} can recover the Ground Truth (GT) DAG given only observational and interventional data. Given observational data, recovery is possible up to a Markov Equivalence Class (MEC). 

We first generate a random ER \cite{er} DAG — with weighted adjacency matrix $W_{GT}$ with sparsity pattern $G_{GT}$, the adjacency matrix — and consider this the ground truth and set the noise covariance to be $\Sigma_{GT} = \sigma_{GT}^2 I$, since we have an isotropic Gaussian assumption. In our experiments, $\sigma_{GT}$ is usually set to $0.1$.

\textbf{Data generation of the true causal variables}: The data generation is done by an ancestral sampling process, compactly given by $z = W^Tz + \epsilon$. One could also get interventional data from this setup: \textbf{(i)} Choose the node set $N$ being intervened upon, \textbf{(ii)} for every node in $N$, zero out the particular column in $W$ to get the mutated DAG $\tilde{W}$ and \textbf{(iii)} perform ancestral sampling using $z = \tilde{W}^Tz + \epsilon$. This process is repeated multiple times to get $n$ samples of $z, $ and we will then organize these samples into a $(n, d)$ matrix and term it $z_{GT}$, or simply, $z$ \footnote{These variables will also be referred to, at times, as samples of true causal variables}. 

\textbf{Generating the true high dimensional data}: We assume the observed low level data is a linear projection of the causal variables given by $X_{GT} = z_{GT}P'$ , where $P' \in \mathbb{R}^{d \times D}$ is a projection matrix and we have $D >> d$ in the real world. For most of the upcoming experiments, we will study recovery in the (simpler) limiting case where $d$ equal to $D$. This work studies recovery in latent space in a case where direct access to true causal variables is not given, in contrast to existing structure learning works.

\section{Decoder BCD}
BCD Nets performs Bayesian inference over $P, L, \Sigma$ given samples of true causal variables to best explain the data by training an ELBO loss (see eq. \ref{bcd_elbo}). Decoder BCD tries to infer the decoder parameters in addition to inferring $P, L, \Sigma$. It is trained over $X_{GT}$ instead of over $z_{GT}$ as in BCD Nets. Thus, we relax the assumption that we have access to samples of true causal variables. We can only access the high dimensional data that has to be explained by $(Z, G)$ and we have to fit a structure, $\hat{G}$, and estimate edge weights, $\hat{W}$, to fit our best guess of the causal variables, $\hat{z}$. Algorithm \ref{alg:cap} summarizes the inference mechanism of Decoder BCD. A diagrammatic overview is given in Figure \ref{model_img}.

\begin{algorithm} 
\caption{Decoder BCD for causal discovery from high dimensional data} \label{alg:cap}
    1. Initialize random distributions for $P$, $L$, $\Sigma$ \\
    2. For \textbf{train\_steps}: \\
    \hspace{0.5cm} (i) Sample $\hat{P}, \hat{L}, \hat{\Sigma} \sim q_{\phi}(P, L, \Sigma)$ \\
    \hspace{0.5cm} (ii) $\hat{W} = (\hat{P}\hat{L}\hat{P^T})^T$ \\
    \hspace{0.5cm} (iii) Perform ancestral sampling: $\hat{z} = \hat{W}^T\hat{z} + \epsilon$; $\epsilon \sim \mathcal{N}(0, \hat{\sigma}^2)$ and $\hat{\Sigma} = \hat{\sigma}^2I$ \\
    \hspace{0.5cm} (iv) Decode $\hat{z}$ to obtain $\hat{X}$ \\
    (v) Update parameters of the distribution $P(P, L, \Sigma)$ with loss as $MSE(X, \hat{X})$ \\
    (vi) For supervised experiments, add an additional KL loss between true and posterior observational joint: $KL(q(z_1, ...z_d) || p(z_1, ...z_d))$
\end{algorithm}

\section{Experiments and Findings}

For all our experiments, for simplicity, we will stick to just learning the decoder and inferring the edge matrix $L$ since this makes the optimization simpler for our studies. Such an assumption of fixing the permutation $P$ to the GT, and thereby, the node orderings, is not unreasonable \cite{graphvae}. Here, the focus is solely on inferring the edges $L$ in latent space whilst learning a decoder. In all experiments, we train the model for $5000$ steps across $20$ random seeds, with a learning rate of 0.002 on ER-2 DAGs. We consider the case of the higher dimensional data being $D=10$ dimensions that is generated by data from a $d=6$ node underlying SCM.

\textbf{Metrics}: In our experiments, we refer to the expected Structural Hamming Distance across 64 samples of the inferred DAG as SHD, $MSE(L, \hat{L})$ is the MSE between predicted $\hat{L}$ and $L_{GT}$, AUROC (a value of 0.5 denotes a random baseline with null edges), and $KL(\text{true}\ ||\ \text{learned})$ refers to the KL divergence between the posterior observational joint and the GT observational joint distributions.

\subsection{Learning edge matrix $L$ with supervision}
For the supervised experiments, we add an additional KL loss on the inferred posterior observational joint $q(z_1,...z_d)$ and the prior observational GT joint distribution, $p(z_1,...z_d) \sim \mathcal{N}(\mu_z, \Sigma_z)$, where $\Sigma_z$ is calculated with $W_{GT}$ instead of with $\hat{W}$. The estimation of the prior and posterior observational joint distribution is detailed in \ref{a1}.

\begin{figure}[h]
\includegraphics[width=8.5cm]{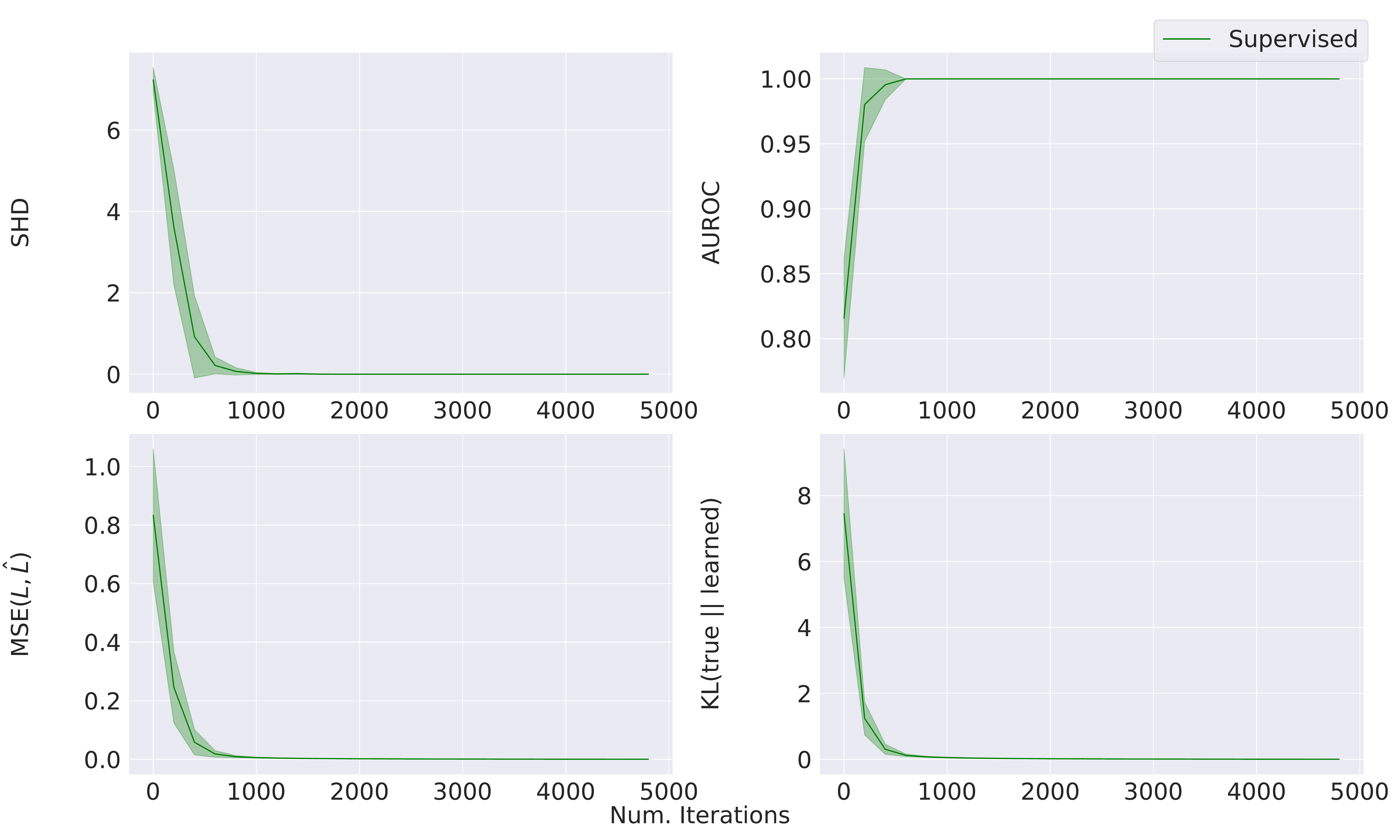} 
\caption{Supervised learning of $L$ on 600 observational data points with $d=6, D=10$
\label{fig1}}
\end{figure}

\textbf{Finding 1}: From figure \ref{fig1}, we can see the expected SHD approaching 0 as $KL(\text{true}\ ||\ \text{learned})$ and $MSE(L, \hat{L})$ approach 0. Using a KL over the observational joint distribution results in complete graph recovery in the supervised case. This is expected since we provide a mild signal for the model to uncover the true causal variables. Instead of providing prior over the samples of true causal variables, we use the true observational joint distribution as a signal.  

\subsection{Unsupervised learning of a single edge weight}
For this setting, instead of inferring the whole lower triangular edge matrix $L$, we infer only the last edge at position $(d, d-1)$. The other elements of the matrix are fixed to the GT and we observe graph recovery in this case. 
This subsection is split into two parts to study recovery with (i) observational data and (ii) a mix of observational and interventional data, to analyze the effects of interventional data.

\begin{figure}[h]
\includegraphics[width=8.5cm]{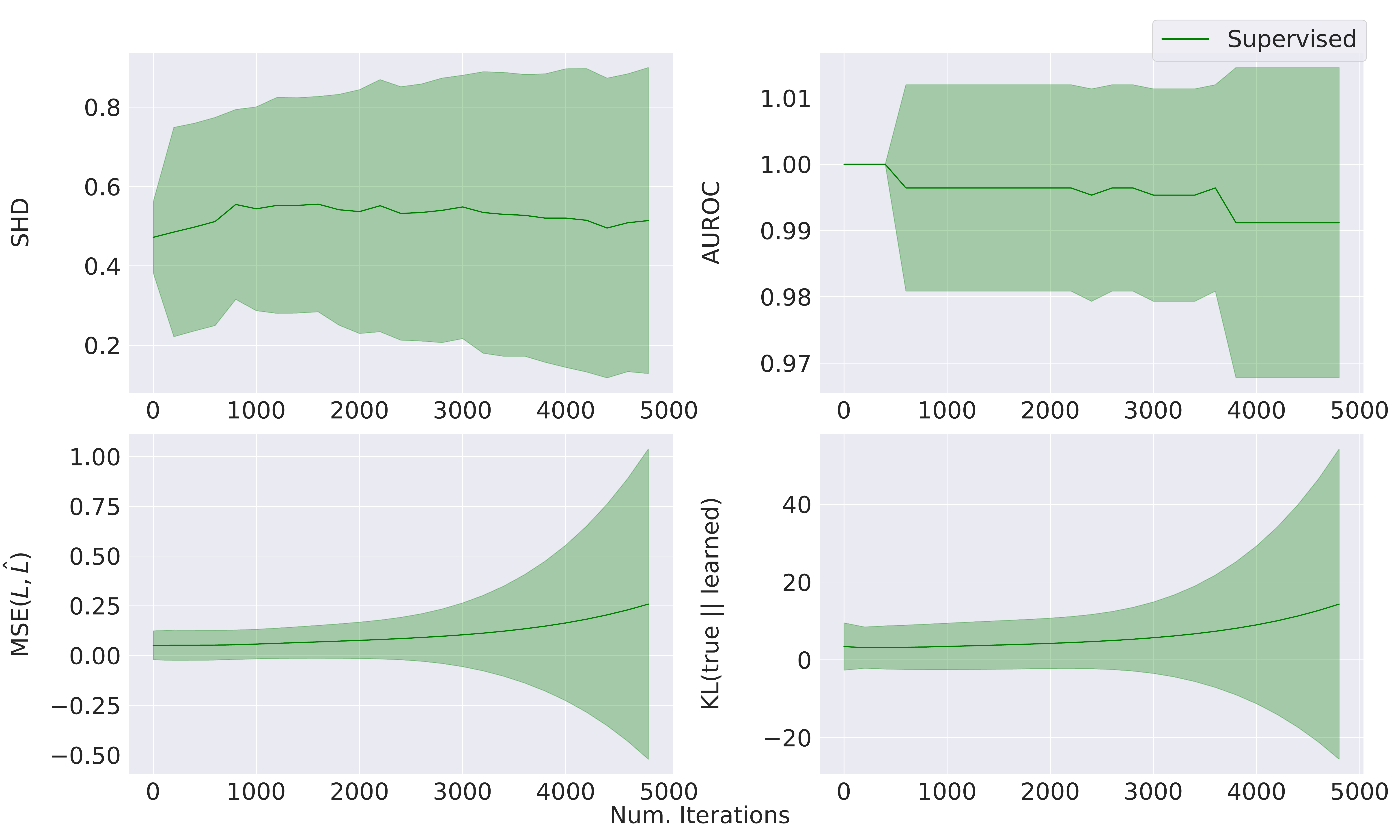} 
\caption{Unsupervised learning of a single edge weight with 1800 observational data points \label{fig2}}
\end{figure}

We use 1800 observational data points for case (i) and 1800 data points (50-50 split of observational and interventional data) for case (ii). The interventional data generation process for single node and multi node interventions is detailed in \ref{a2}. The interventional values are fixed to 100.

\begin{figure}[h]
\includegraphics[width=8.5cm]{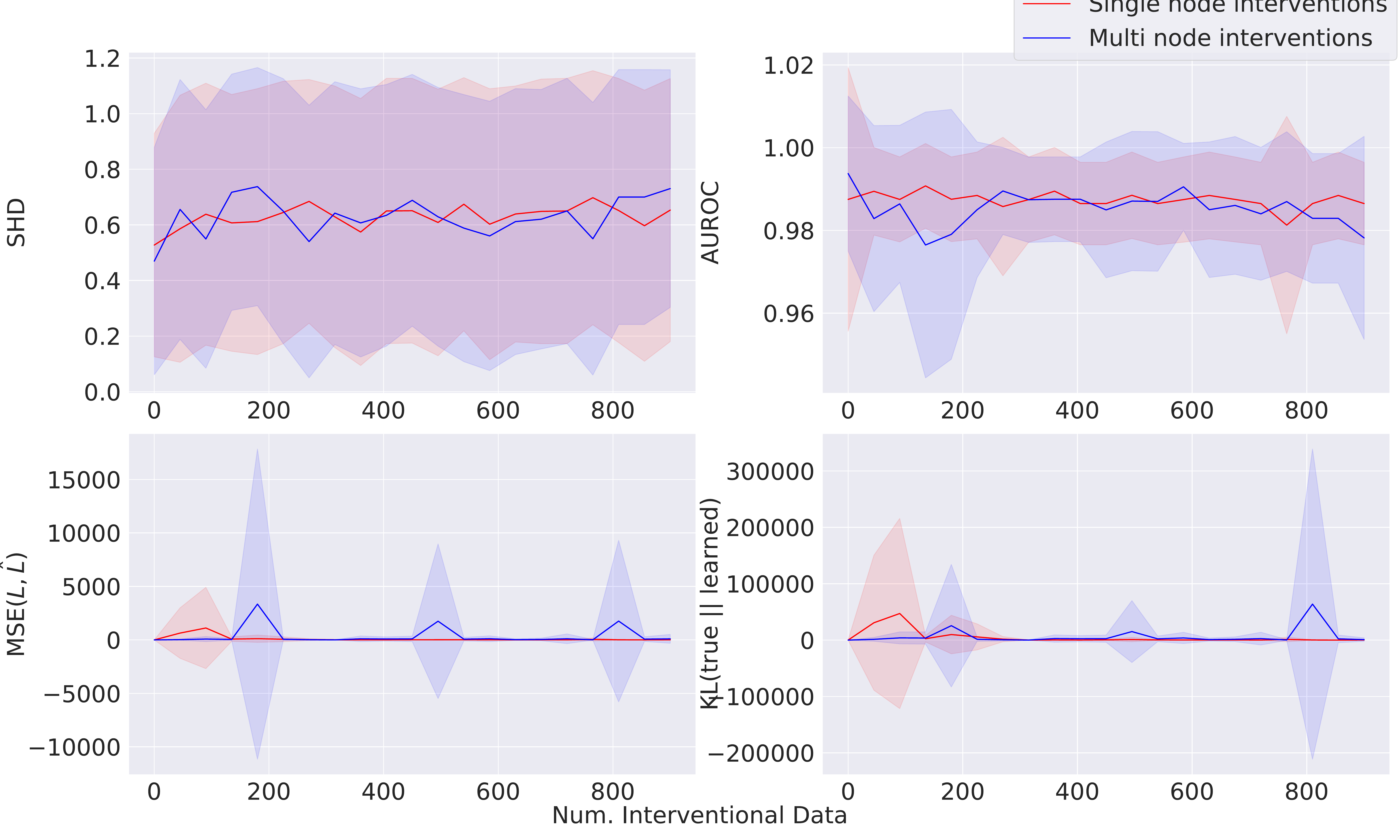} 
\caption{Unsupervised learning of a single edge weight with 900 observational and 900 interventional data points \label{fig2.2}}
\end{figure}

\textbf{Finding 2}: Figures \ref{fig2} and \ref{fig2.2} reveal that using observational and/or interventional data with single node or multi node interventions with fixed intervention values is not sufficient to learn to orient a single edge in the unsupervised case.

\subsection{Unsupervised learning of edge weight $L$}
\begin{figure}[h]
\includegraphics[width=8.5cm]{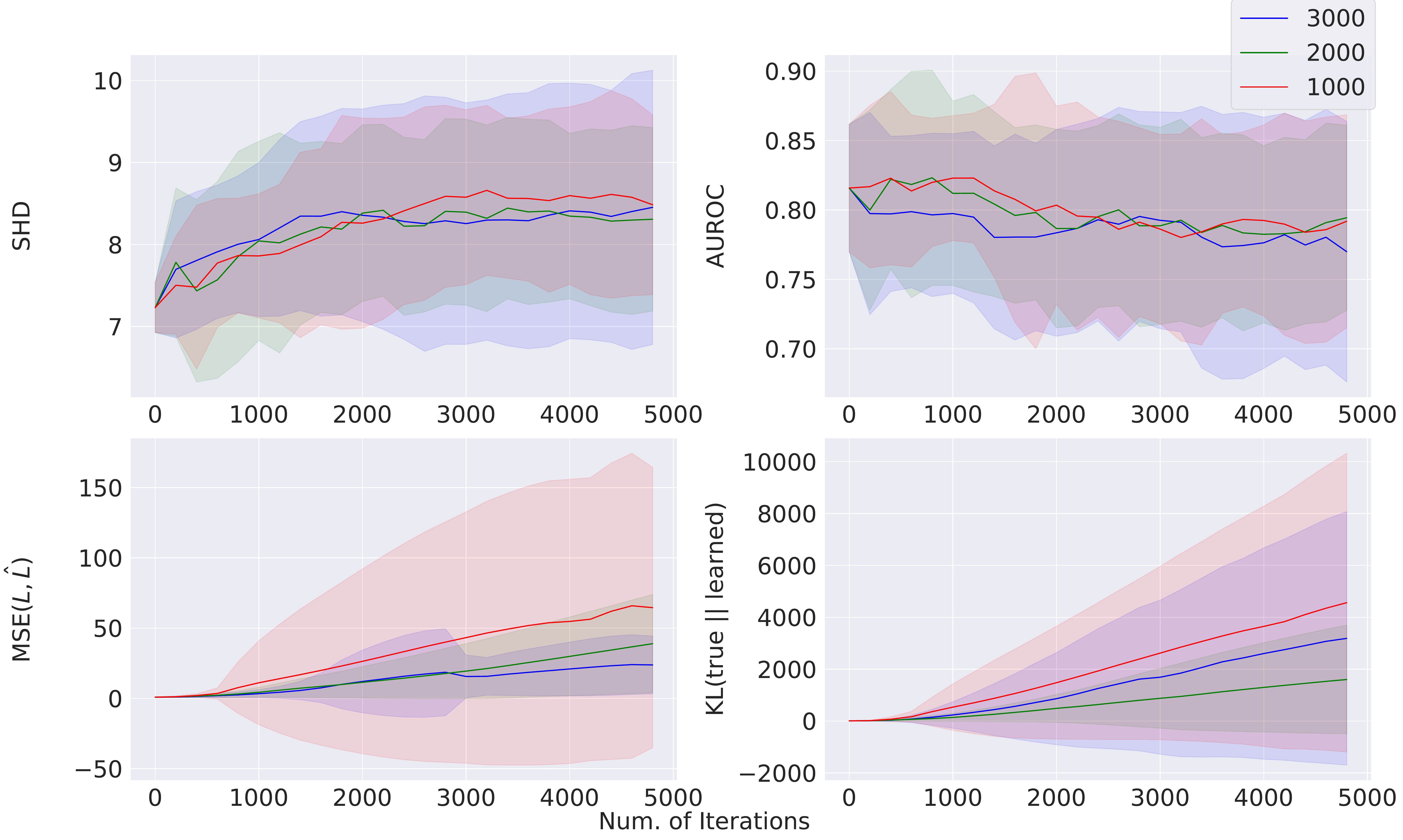} 
\caption{Unsupervised learning of the edge weight matrix with various amounts of observational data \label{fig3.1}}
\end{figure}

In this experiment and in the next, we explore the problem of learning the entire lower triangular edge matrix $L$ in an unsupervised setting. First, we consider the learning problem with various amounts of observational data to analyze its effect on edge recovery in the latent space, which is shown in figure \ref{fig3.1}.

Figure \ref{fig3.1} shows that all metrics diverge with training but there is no trend with respect to the amount of observational data that the model is given.

\begin{figure}[h]
\includegraphics[width=8.5cm]{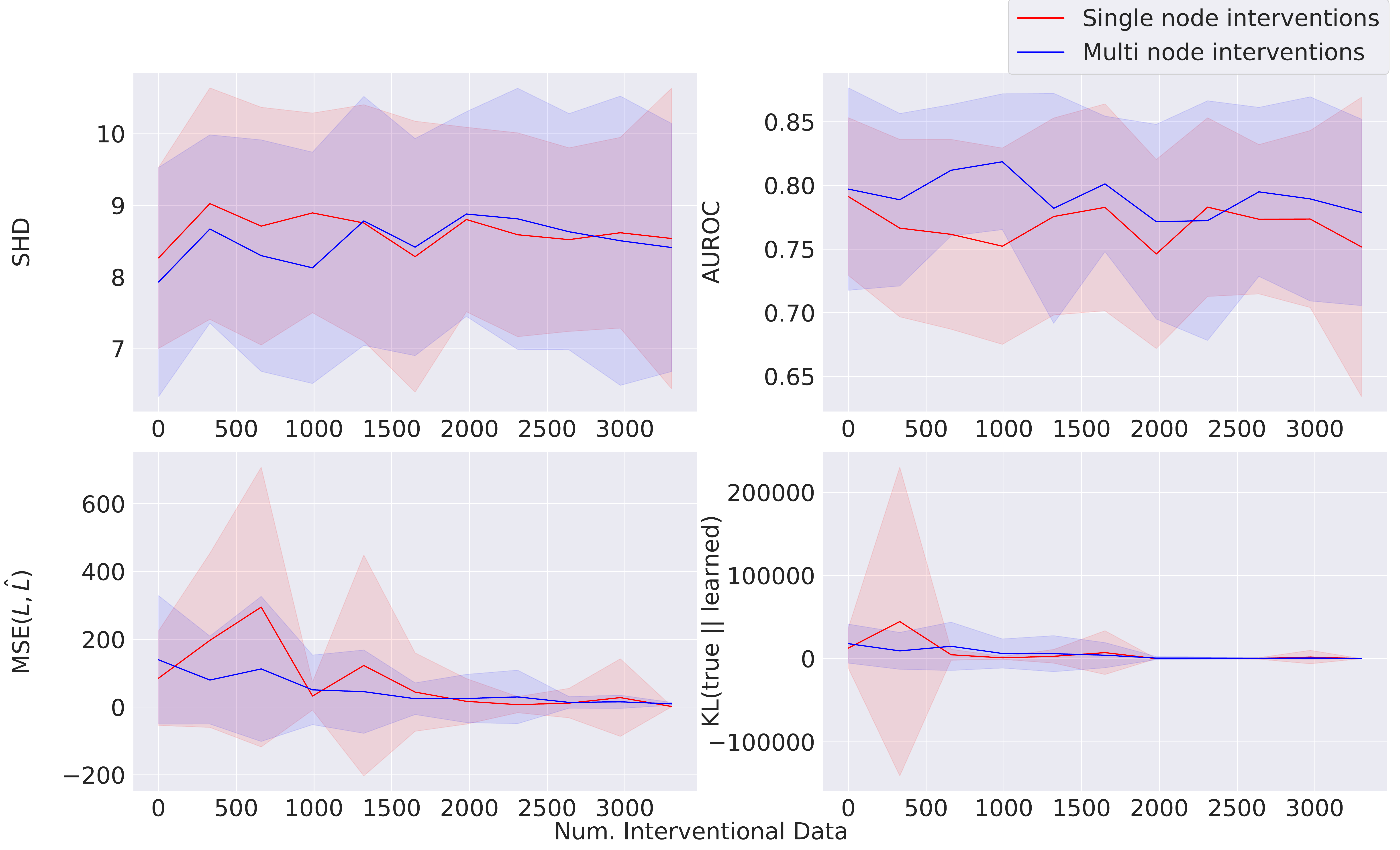} 
\caption{Unsupervised learning of edge weights with random single node and multi node interventions \label{fig4.1}}
\end{figure}

We now consider learning $L$ using interventional targets as labels to learn the structure in the latent space with a mix of 300 observational data points and 3300 interventional data points. For this experiment, we retrain Decoder BCD multiple times from scratch — each time with the same 300 observational data points but with  more interventional data than the previous run. This helps us understand the usefulness of interventional data for graph recovery. This result is illustrated in figure \ref{fig4.1}. For the interventional data points, we chose to use a fixed intervention value of 100.0. The reason for this particular value was that we had to choose a value that is far from 0 -- the mean of all the nodes in the causal graph. Note that in a linear Gaussian additive noise SCM, if one has 0 mean of the error variables $\epsilon$, then all nodes in the graph have 0 mean.

\begin{figure}[h]
\includegraphics[width=8.5cm]{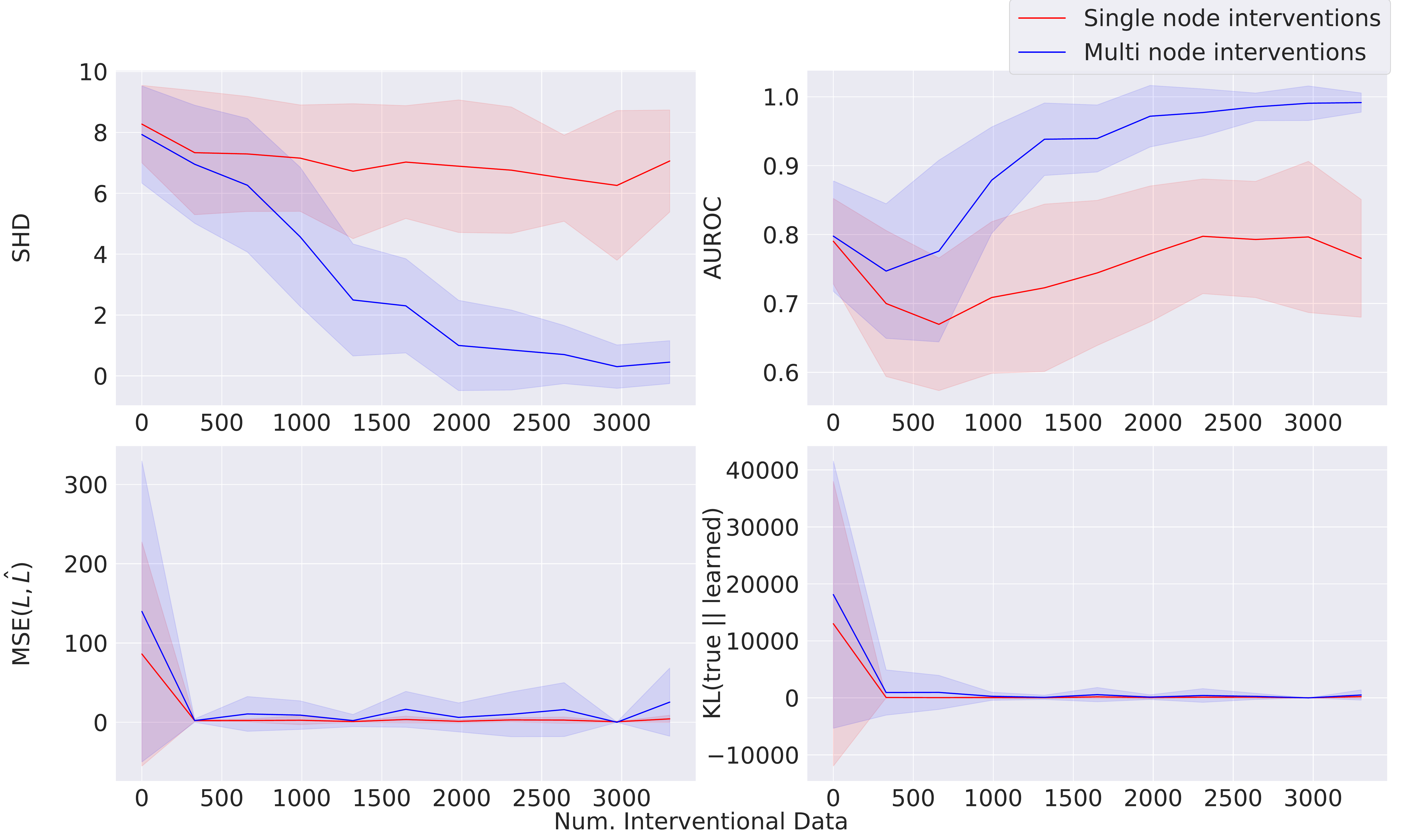} 
\caption{Unsupervised learning of edge weights with uniform single node and multi node interventions \label{fig4.3}}
\end{figure}

Finally, other than learning from randomly (single or multi) intervened nodes, we also performed an experiment to observe the effect of randomly chosen interventional values, rather than a fixed interventional value. Thus, instead of an intervention value of 100, we randomly sampled interventional values in $\text{Uniform}(-10.0, 10.0)$ for each data point and repeated our previous experiment for the same amount of observational (300) and interventional (3300) data points. We summarize our findings for this experiment in figure \ref{fig4.3}.

\textbf{Finding 4}: Figures \ref{fig4.1} and \ref{fig4.3} show that both single node and multi node interventions help in recovering the edge weights, measured across all the 4 metrics. However, multi node interventions with uniformly sampled intervention values results in the better inference of the structure and parameters of the latent SCM.

\section{Related Works}

To address the challenges of causal discovery, a variety of methods have been proposed. Some of these methods are based on structure learning using observational data and some take in to account interventional data \cite{dibs, ncmfai, ncmfui, dcdi}. 

There has been an increasing focus on Bayesian structure learning \cite{daggnn, vcn, dibs, bcdnets, dag_gflow} to quantify epistemic uncertainty that is crucial for reinforcement learning and active learning settings. \citet{dds} follow almost exactly the same approach as \citet{bcdnets} except that they operate on nonlinear Gaussian SCM instead of a linear one. There also exist many maximum likelihood based methods, one such example is \cite{graphvae}. It is one of the few works that learn a structure in the latent space but they do not operate in a causality-based or SCM framework.

Markov Chain Monte Carlo (MCMC) is a popular technique for sampling from complex high dimensional probability distributions, such as the posterior distribution of DAGs. \cite{madigan1995bayesian} uses Metropolis-Hastings \cite{mhalgo} to predict the posterior distribution through Markov space to perform single edge addition or deletion. \cite{eaton2007exact} propose a hybrid MCMC algorithm that uses an exact score based algorithm. \cite{kuipers2022efficient} and \cite{viinikka2020towards} use more efficient MCMC samplers. \cite{dag_gflow} uses a novel class of probabilistic models, GFlowNets \cite{gflow}, which model distribution over discrete entities like DAGs to approximate the posterior in place of MCMC algorithms. Section \ref{mrw} discusses related work in more detail. Finally, our work involved randomly selecting nodes to intervene on while also randomly selecting the values for the interventions. \cite{iwah} uses a mutual information objective to learn where (which nodes) and how (with what values) to perform interventions in an active learning scenario to recover the edges more efficiently. However, unlike ours, their SCM is not in the latent space.

\section{Conclusion}
In this work, we introduced our latent variable model, Decoder BCD, and studied the causal representation learning problem. We explored the cases where edge recovery fails -- learning to orient a single edge or learning with only observational data. To address this, we propose using interventional targets as labels to allow recovery  of edges and edge weights in an SCM. Our experiments show that this is a promising direction for the unsupervised Bayesian causal discovery in latent space. However, our hypothesis of the observed data having a latent linear SCM and linear projection of the latent causal variables to higher dimension is a limitation (refer \ref{limitations}) when it comes to mechanisms in the real world. Future work should explore nonlinear projections of the causal variables as well as nonlinear and non-Gaussian SCMs. Finally, we discuss some key challenges for future study in Appendix \ref{challenges}. 

\section*{Acknowledgements}

The authors are grateful to Nan Rosemary Ke, Anirudh Goyal, Tristan Deleu and Sébastien Lachapelle for fruitful discussions and feedback. The authors are also thankful to Compute Canada and CIFAR for the compute and funding that made this work possible. 
\bibliography{refs}

\begin{thebibliography}{33}
\providecommand{\natexlab}[1]{#1}
\providecommand{\url}[1]{\texttt{#1}}
\expandafter\ifx\csname urlstyle\endcsname\relax
  \providecommand{\doi}[1]{doi: #1}\else
  \providecommand{\doi}{doi: \begingroup \urlstyle{rm}\Url}\fi

\bibitem[Annadani et~al.(2021)Annadani, Rothfuss, Lacoste, Scherrer, Goyal,
  Bengio, and Bauer]{vcn}
Annadani, Y., Rothfuss, J., Lacoste, A., Scherrer, N., Goyal, A., Bengio, Y.,
  and Bauer, S.
\newblock Variational causal networks: Approximate bayesian inference over
  causal structures, 2021.
\newblock URL \url{https://arxiv.org/abs/2106.07635}.

\bibitem[Bengio et~al.(2021)Bengio, Jain, Korablyov, Precup, and Bengio]{gflow}
Bengio, E., Jain, M., Korablyov, M., Precup, D., and Bengio, Y.
\newblock Flow network based generative models for non-iterative diverse
  candidate generation, 2021.
\newblock URL \url{https://arxiv.org/abs/2106.04399}.

\bibitem[Brouillard et~al.(2020)Brouillard, Lachapelle, Lacoste,
  Lacoste-Julien, and Drouin]{dcdi}
Brouillard, P., Lachapelle, S., Lacoste, A., Lacoste-Julien, S., and Drouin, A.
\newblock Differentiable causal discovery from interventional data, 2020.
\newblock URL \url{https://arxiv.org/abs/2007.01754}.

\bibitem[Charpentier et~al.(2022)Charpentier, Kibler, and Günnemann]{dds}
Charpentier, B., Kibler, S., and Günnemann, S.
\newblock Differentiable dag sampling, 2022.

\bibitem[Cundy et~al.(2021)Cundy, Grover, and Ermon]{bcdnets}
Cundy, C., Grover, A., and Ermon, S.
\newblock Bcd nets: Scalable variational approaches for bayesian causal
  discovery, 2021.
\newblock URL \url{https://arxiv.org/abs/2112.02761}.

\bibitem[Deleu et~al.(2022)Deleu, Góis, Emezue, Rankawat, Lacoste-Julien,
  Bauer, and Bengio]{dag_gflow}
Deleu, T., Góis, A., Emezue, C., Rankawat, M., Lacoste-Julien, S., Bauer, S.,
  and Bengio, Y.
\newblock Bayesian structure learning with generative flow networks, 2022.
\newblock URL \url{https://arxiv.org/abs/2202.13903}.

\bibitem[Eaton \& Murphy(2007)Eaton and Murphy]{eaton2007exact}
Eaton, D. and Murphy, K.
\newblock Exact bayesian structure learning from uncertain interventions.
\newblock In Meila, M. and Shen, X. (eds.), \emph{Proceedings of the Eleventh
  International Conference on Artificial Intelligence and Statistics}, volume~2
  of \emph{Proceedings of Machine Learning Research}, pp.\  107--114, San Juan,
  Puerto Rico, 21--24 Mar 2007. PMLR.
\newblock URL \url{https://proceedings.mlr.press/v2/eaton07a.html}.

\bibitem[Erdos et~al.(1960)Erdos, R{\'e}nyi, et~al.]{er}
Erdos, P., R{\'e}nyi, A., et~al.
\newblock On the evolution of random graphs.
\newblock \emph{Publ. Math. Inst. Hung. Acad. Sci}, 5\penalty0 (1):\penalty0
  17--60, 1960.

\bibitem[Geiger \& Heckerman(1994)Geiger and Heckerman]{geiger1994learning}
Geiger, D. and Heckerman, D.
\newblock Learning gaussian networks.
\newblock In \emph{Uncertainty Proceedings 1994}, pp.\  235--243. Elsevier,
  1994.

\bibitem[Ghoshal \& Honorio(2017)Ghoshal and Honorio]{linearscmpolytime}
Ghoshal, A. and Honorio, J.
\newblock Learning linear structural equation models in polynomial time and
  sample complexity.
\newblock 2017.
\newblock \doi{10.48550/ARXIV.1707.04673}.
\newblock URL \url{https://arxiv.org/abs/1707.04673}.

\bibitem[He et~al.(2019)He, Gong, Marino, Mori, and Lehrmann]{graphvae}
He, J., Gong, Y., Marino, J., Mori, G., and Lehrmann, A.
\newblock Variational autoencoders with jointly optimized latent dependency
  structure.
\newblock In \emph{International Conference on Learning Representations}, 2019.
\newblock URL \url{https://openreview.net/forum?id=SJgsCjCqt7}.

\bibitem[Jang et~al.(2016)Jang, Gu, and Poole]{gs}
Jang, E., Gu, S., and Poole, B.
\newblock Categorical reparameterization with gumbel-softmax, 2016.
\newblock URL \url{https://arxiv.org/abs/1611.01144}.

\bibitem[Ke et~al.(2019)Ke, Bilaniuk, Goyal, Bauer, Larochelle, Schölkopf,
  Mozer, Pal, and Bengio]{ncmfui}
Ke, N.~R., Bilaniuk, O., Goyal, A., Bauer, S., Larochelle, H., Schölkopf, B.,
  Mozer, M.~C., Pal, C., and Bengio, Y.
\newblock Learning neural causal models from unknown interventions, 2019.
\newblock URL \url{https://arxiv.org/abs/1910.01075}.

\bibitem[Kuipers et~al.(2022)Kuipers, Suter, and Moffa]{kuipers2022efficient}
Kuipers, J., Suter, P., and Moffa, G.
\newblock Efficient sampling and structure learning of bayesian networks.
\newblock \emph{Journal of Computational and Graphical Statistics}, pp.\
  1--12, 2022.

\bibitem[Loh \& B{\"u}hlmann(2014)Loh and B{\"u}hlmann]{highdimcausal}
Loh, P.-L. and B{\"u}hlmann, P.
\newblock High-dimensional learning of linear causal networks via inverse
  covariance estimation.
\newblock \emph{The Journal of Machine Learning Research}, 15\penalty0
  (1):\penalty0 3065--3105, 2014.

\bibitem[Lorch et~al.(2021)Lorch, Rothfuss, Sch\"{o}lkopf, and Krause]{dibs}
Lorch, L., Rothfuss, J., Sch\"{o}lkopf, B., and Krause, A.
\newblock Dibs: Differentiable bayesian structure learning.
\newblock In Ranzato, M., Beygelzimer, A., Dauphin, Y., Liang, P., and Vaughan,
  J.~W. (eds.), \emph{Advances in Neural Information Processing Systems},
  volume~34, pp.\  24111--24123. Curran Associates, Inc., 2021.
\newblock URL
  \url{https://proceedings.neurips.cc/paper/2021/file/ca6ab34959489659f8c3776aaf1f8efd-Paper.pdf}.

\bibitem[Madigan et~al.(1995)Madigan, York, and Allard]{madigan1995bayesian}
Madigan, D., York, J., and Allard, D.
\newblock Bayesian graphical models for discrete data.
\newblock \emph{International Statistical Review/Revue Internationale de
  Statistique}, pp.\  215--232, 1995.

\bibitem[Metropolis et~al.(1953)Metropolis, Rosenbluth, Rosenbluth, Teller, and
  Teller]{mhalgo}
Metropolis, N., Rosenbluth, A.~W., Rosenbluth, M.~N., Teller, A.~H., and
  Teller, E.
\newblock Equation of state calculations by fast computing machines.
\newblock \emph{The Journal of Chemical Physics}, 21\penalty0 (6):\penalty0
  1087--1092, 1953.
\newblock \doi{10.1063/1.1699114}.
\newblock URL \url{http://link.aip.org/link/?JCP/21/1087/1}.

\bibitem[Mooij et~al.(2016)Mooij, Peters, Janzing, Zscheischler, and
  Sch{\"o}lkopf]{mooij2016distinguishing}
Mooij, J.~M., Peters, J., Janzing, D., Zscheischler, J., and Sch{\"o}lkopf, B.
\newblock Distinguishing cause from effect using observational data: methods
  and benchmarks.
\newblock \emph{The Journal of Machine Learning Research}, 17\penalty0
  (1):\penalty0 1103--1204, 2016.

\bibitem[Ng et~al.(2020)Ng, Ghassami, and Zhang]{golem}
Ng, I., Ghassami, A., and Zhang, K.
\newblock On the role of sparsity and dag constraints for learning linear dags.
\newblock In Larochelle, H., Ranzato, M., Hadsell, R., Balcan, M., and Lin, H.
  (eds.), \emph{Advances in Neural Information Processing Systems}, volume~33,
  pp.\  17943--17954. Curran Associates, Inc., 2020.
\newblock URL
  \url{https://proceedings.neurips.cc/paper/2020/file/d04d42cdf14579cd294e5079e0745411-Paper.pdf}.

\bibitem[Pamfil et~al.(2020)Pamfil, Sriwattanaworachai, Desai, Pilgerstorfer,
  Beaumont, Georgatzis, and Aragam]{dynotears}
Pamfil, R., Sriwattanaworachai, N., Desai, S., Pilgerstorfer, P., Beaumont, P.,
  Georgatzis, K., and Aragam, B.
\newblock Dynotears: Structure learning from time-series data, 2020.
\newblock URL \url{https://arxiv.org/abs/2002.00498}.

\bibitem[Pearl(2009)]{pearl}
Pearl, J.
\newblock \emph{Causality}.
\newblock Cambridge university press, 2009.

\bibitem[Peters \& Bühlmann(2013)Peters and Bühlmann]{peters_eq_noi_var}
Peters, J. and Bühlmann, P.
\newblock Identifiability of gaussian structural equation models with equal
  error variances.
\newblock \emph{Biometrika}, 101\penalty0 (1):\penalty0 219--228, nov 2013.
\newblock \doi{10.1093/biomet/ast043}.
\newblock URL \url{https://doi.org/10.1093%2Fbiomet%2Fast043}.

\bibitem[Peters et~al.(2017)Peters, Janzing, and
  Sch{\"o}lkopf]{peters2017elements}
Peters, J., Janzing, D., and Sch{\"o}lkopf, B.
\newblock \emph{Elements of causal inference: foundations and learning
  algorithms}.
\newblock The MIT Press, 2017.

\bibitem[Scherrer et~al.(2021)Scherrer, Bilaniuk, Annadani, Goyal, Schwab,
  Schölkopf, Mozer, Bengio, Bauer, and Ke]{ncmfai}
Scherrer, N., Bilaniuk, O., Annadani, Y., Goyal, A., Schwab, P., Schölkopf,
  B., Mozer, M.~C., Bengio, Y., Bauer, S., and Ke, N.~R.
\newblock Learning neural causal models with active interventions, 2021.
\newblock URL \url{https://arxiv.org/abs/2109.02429}.

\bibitem[Schölkopf et~al.(2021)Schölkopf, Locatello, Bauer, Ke, Kalchbrenner,
  Goyal, and Bengio]{tcrl}
Schölkopf, B., Locatello, F., Bauer, S., Ke, N.~R., Kalchbrenner, N., Goyal,
  A., and Bengio, Y.
\newblock Towards causal representation learning, 2021.
\newblock URL \url{https://arxiv.org/abs/2102.11107}.

\bibitem[Shah \& Peters(2020)Shah and Peters]{Shah_2020}
Shah, R.~D. and Peters, J.
\newblock The hardness of conditional independence testing and the generalised
  covariance measure.
\newblock \emph{The Annals of Statistics}, 48\penalty0 (3), jun 2020.
\newblock \doi{10.1214/19-aos1857}.
\newblock URL \url{https://doi.org/10.1214%2F19-aos1857}.

\bibitem[Shimizu et~al.(2006)Shimizu, Hoyer, Hyv{\"a}rinen, Kerminen, and
  Jordan]{lingam}
Shimizu, S., Hoyer, P.~O., Hyv{\"a}rinen, A., Kerminen, A., and Jordan, M.
\newblock A linear non-gaussian acyclic model for causal discovery.
\newblock \emph{Journal of Machine Learning Research}, 7\penalty0 (10), 2006.

\bibitem[Tigas et~al.(2022)Tigas, Annadani, Jesson, Schölkopf, Gal, and
  Bauer]{iwah}
Tigas, P., Annadani, Y., Jesson, A., Schölkopf, B., Gal, Y., and Bauer, S.
\newblock Interventions, where and how? experimental design for causal models
  at scale, 2022.
\newblock URL \url{https://arxiv.org/abs/2203.02016}.

\bibitem[Viinikka et~al.(2020)Viinikka, Hyttinen, Pensar, and
  Koivisto]{viinikka2020towards}
Viinikka, J., Hyttinen, A., Pensar, J., and Koivisto, M.
\newblock Towards scalable bayesian learning of causal dags.
\newblock In Larochelle, H., Ranzato, M., Hadsell, R., Balcan, M., and Lin, H.
  (eds.), \emph{Advances in Neural Information Processing Systems}, volume~33,
  pp.\  6584--6594. Curran Associates, Inc., 2020.
\newblock URL
  \url{https://proceedings.neurips.cc/paper/2020/file/48f7d3043bc03e6c48a6f0ebc0f258a8-Paper.pdf}.

\bibitem[Yang et~al.(2022)Yang, Liu, Chen, Shen, Hao, and Wang]{causalvae}
Yang, M., Liu, F., Chen, Z., Shen, X., Hao, J., and Wang, J.
\newblock Causalvae: Structured causal disentanglement in variational
  autoencoder, 2022.

\bibitem[Yu et~al.(2019)Yu, Chen, Gao, and Yu]{daggnn}
Yu, Y., Chen, J., Gao, T., and Yu, M.
\newblock Dag-gnn: Dag structure learning with graph neural networks, 2019.
\newblock URL \url{https://arxiv.org/abs/1904.10098}.

\bibitem[Zheng et~al.(2018)Zheng, Aragam, Ravikumar, and Xing]{notears}
Zheng, X., Aragam, B., Ravikumar, P.~K., and Xing, E.~P.
\newblock Dags with no tears: Continuous optimization for structure learning.
\newblock In Bengio, S., Wallach, H., Larochelle, H., Grauman, K.,
  Cesa-Bianchi, N., and Garnett, R. (eds.), \emph{Advances in Neural
  Information Processing Systems}, volume~31. Curran Associates, Inc., 2018.
\newblock URL
  \url{https://proceedings.neurips.cc/paper/2018/file/e347c51419ffb23ca3fd5050202f9c3d-Paper.pdf}.

\end{thebibliography}
\bibliographystyle{icml2022}

\newpage
\appendix
\onecolumn

\section{Appendix}

\subsection{KL Loss for the mildly supervised experiments} \label{a1}
 Given $\hat{W} = ({P\hat{L}P^T})^T$, from inferred $\hat{L}$, one can obtain the mean and covariance of the observational joint distribution $q(z_1,...z_d)$ as follows:

\begin{equation}
    z = \hat{W}^Tz + \epsilon ;\hspace{0.3cm} \epsilon \sim \mathcal{N}(0, \sigma) 
\end{equation}
\begin{equation}
    z = (I - \hat{W})^{-T} \epsilon
\end{equation}
\begin{equation}
    q(z_1,...z_d) \sim \mathcal{N}(\hat{\mu_z}, \hat{\Sigma_z})
\end{equation}
\begin{equation} \label{estobs}
    \hat{\mu_z} = 0 \hspace{0.1cm} and \hspace{0.1cm} \hat{\Sigma_z} = (I - \hat{W})^{-T} \Sigma (I - \hat{W})^{-1}
\end{equation}

To estimate the prior GT observational joint distribution, one would use $W$ in place of $\hat{W}$ in equation \ref{estobs}.

\subsection{Generating interventional data for experiments that use single node and multi node interventions} \label{a2}

Suppose we have to generate $i$ interventional data points. We split the data generation process into $s=20$ sets, each set generating $i/s$ interventional data points. \textit{For single node interventions}, we randomly choose a node and sample $i/s$ data points. The process is repeated $s$ times randomly to generate the $i$ data points. \textit{For multi node interventions}, we randomly choose a number in $[2, d]$ to decide on the number of nodes to intervene on (call this $x$). We then choose $x$ nodes without replacement and perform the interventions on these nodes and sample $i/s$ data points. The process is repeated $s$ times randomly to generate the $i$ data points.

\subsection{More Related Work} \label{mrw}

Since discrete optimization is hard and often involves enumeration of possible structures, the super-exponential nature of structure learning has resulted in the community resorting to relaxing the discrete optimization problem into a continuous one \cite{dibs, bcdnets, vcn, ncmfai, ncmfui, notears} and learning the parameters using gradient descent. \cite{highdimcausal} propose a scalable, scoring-based DAG learning approach to recover high dimensional, sparse causal graphs in a non-Gaussian setting where only some but not all exogenous noise variables are expected to be non-Gaussian. \cite{linearscmpolytime} learns a linear structural equation model in polynomial time. \cite{ncmfui} learns the causal structure from unknown interventions but operates on the Bernoulli distribution while \cite{ncmfai} is in an active learning framework and the system determines the intervention that will be most useful in gaining knowledge about the graph structure. \cite{causalvae} proposes a variational autoencoder parameterised by exogenous variables to learn causal semantics of the data. Another family of works introduce assumptions to functional and parametric form of the data-generation structure. They exploit symmetries to learn the causal structure \cite{peters2017elements, mooij2016distinguishing}. \\

Approaches to the problem are mostly employ score-based or constraint-based optimization. Most modern methods use some sort of a scoring function to rank estimated structure and use it to rank structures and optimizing for the score is expected to return the ground truth DAG. Popular scoring functions include Bayesian Information Criterion (BIC) and Bayesian Gaussian Equivalent marginal likelihood score \cite{geiger1994learning}. These methods typically use a regularization over the structure to induce sparsity and/or acyclicity. Some methods impose hard constraints as well that ensure the search is done only over the space of DAGs. \cite{Shah_2020} is a constraint based approach that tests for conditional independence.

\subsection{Key challenges for future study} \label{challenges}

One of the most important scientific questions of causal representation learning is regarding the \textbf{relationship between high dimensional, observed variables and the low dimensional, causal variables}: In this work, we perform synthetic data generation of $z$ and project it to higher dimensions by using a random projection matrix $P'$. We begin on the premise that real-world, high dimensional data can be explained by a few causal variables and the inferring these variables and their structure is the problem of causal inference that the brain solves for performing intelligent tasks. Thus, there must exist an operation that maps the low dimensional causal variables to the high dimensional, observed variables (eg. images, videos). For our problem setting, we assume this is true and try to generate high dimensional samples that are "causally consistent" by performing a linear projection $X = zP'$. However, we do not know how this process of projection to higher dimensions might happen in reality. \\

\textbf{What’s the right loss function for unsupervised causal discovery?}
In all experiments, we found that the MSE over high dimensional data, $X$, goes down but this does not necessarily mean that graph recovery in the latent space gets better. Therefore, we need to look for alternative losses with a property such that reduction in loss over $X$ guarantees a better recovery in the latent space (i.e., better graph structure recovery or better estimates of edge weight matrix $L$). Ideally, such a loss should result in a reduction in the KL divergence between the inferred posterior observational joint distribution and the GT observational joint distribution. We propose that this a better metric to measure since in the supervised experiments, getting a low enough value of this metric results in the SHD dropping steeply to 0.

\subsection{Limitations} \label{limitations}
A limitation of this work is that we do not know if it is practical to assume a linear projection -- it is just a formulation that we explore. Additionally, if it \textit{is} a linear operation, are there any properties that the projection matrix $P'$ must hold to maintain this "causal consistency" in higher dimensions? If $P'$ needs to hold some properties for causal inference to be performed from high dimensions, \textit{what} exactly are these properties? It is easy to see that a random projection matrix (which transforms a $d$-dimensional vector to $D$-dimensional vector) can be random enough to completely destroy the encoded information due to the causal generation process that occurred in the lower dimensions, and thus the high dimensional data could no longer be "causally consistent" for us to perform inference. And finally, one needs to focus on the question of whether the projection operation could be nonlinear.
\end{document}